# Isotopic envelope identification by analysis of the spatial distribution of components in MALDI-MSI data

Anna Glodek[1], Joanna Polańska[1], Marta Gawin[2]

[1] Department of Data Science and Engineering, The Silesian University of Technology, Gliwice, Poland
[2] Maria Sklodowska-Curie National Research Institute of Oncology Gliwice Branch

```
{anna.glodek, joanna.polanska}@polsl.pl
         marta.gawin@io.gliwice.pl
```

**INTRODUCTION**

One of the most crucial biomolecules is proteins since they play a key role in several biological processes; they are functional or structural elements of the cells [1], and many o them act as clinical biomarkers [2]. Due to the different lengths of amino acid chains that proteins are built of, they are divided into three groups: 2-20 amino acids are the oligopeptides, 20-50 are peptides, and over 50 amino acids are proteins [3-4]. Proteins are often digested into peptides (e.g., by trypsin) in order to meet a condition of small enough size to perform a mass spectrometry (MS) experiment [5]. In order to understand the nature of the tissue, the molecular features of the tissue should be calculated [6]. To preserve the sample for further analysis, Mass Spectrometry Imaging (MALDI-MSI), which is performed directly on the sample surface, was performed [7]. For MALDI-MSI, the sample has to be prepared in the subsequent steps: sectioned, stained (optional), washed, on-tissue digested and coated with matrix [6, 8]. After that, the sample is loaded into the mass spectrometer. The following steps occur: a laser beam shoots in the matrix layer, which results in ionizing analytes, and a mass spectrum is measured for every spot. With the mass spectrum, it is possible to present given *m/z* signals as spatial distribution molecular maps of intensities. Generally, there is a plethora of algorithms that can identify an isotopic envelope. However, most of them are dedicated to specific types of biomolecules or experiments and are unsuitable for large datasets of MALDI-MSI-driven data. This work proposes a novel approach (*DeisoLAB*) for MALDI-MSI data toward isotopic envelope identification based on the Mamdani-Assilan fuzzy-inference system for preselection and the Naïve Bayes classifier for the final classification. The tests of the proposed method were performed on eight datasets: 4 of them come from fresh frozen tissue material (FF), and the remaining ones are from Formalin-Fixed Paraffin-Embedded method (FFPE) from the head and neck cancer suffered patients.

**MATERIALS AND METHODS**

*Existing approaches.* The main challenge when identifying isotopic envelopes is accurately detecting those that describe the mass spectrum and handling overlapping isotopic envelopes [30]. The two most common approaches are Graph theory-based (GT), which focuses on constructing isotopic-cluster graphs, and an approach based on matching the theoretical isotopic distribution with the experimental one (TvsE) – most of them use the *Averagine* method, which allows for monoisotopic mass determination of the measured isotopic distribution by comparison with a model molecule that has same average molecular mass [53].

*Table 1. Comparison of several deisotoping algorithms.*

| Name | Mass spectrometry technique/proteomic strategy | Biomolecules | Approach |
|---|---|---|---|
| **DeconMSn [52]** | LC-MS/MS | peptides | TvsE |
| **Decon2LS [43]** | LC-MS/MS | proteins, peptides, metabolites | TvsE |
| **FLASHDeconv [46]** | Top-down proteomics | proteins, peptides | TvsE |
| **MS-Deconv [40]** | Tandem mass spectra, Top-down proteomics | proteins, peptides | GT |
| **MS2-Deisotoper [41]** | High resolution bottom-up spectra, MS/MS | peptides | GT |
| **RAPID [49]** | LC-MS/MS | - | TvsE |
| **BPDA [36]** | MALDI-TOF-MS, LC-MS | peptides | other: a Bayesian peptide detection algorithm |
| **Features-Based Deisotoping Method [39]** | Tandem mass spectra, bottom-up spectra | proteins, peptides | GT |
| **THRASH [42]** | MALDI-MS, ESMS | proteins, peptides, DNA, polymers | TvsE |
| **pyOpenMS [44]** | LC-MS/MS | proteins, peptides, metabolites | TvsE |
| **NITPICK [45]** | not limited | not limited | TvsE |
| **mMass [47]** | MALDI-TOF-MS, all others | proteins, peptides | TvsE |
| **iMEF & ProteinGoggle 2.0 [48]** | Tandem mass spectra | proteins, peptides | TvsE |
| **Xtract [50]** | Top-down proteomics | proteins | other |
| **Zscore [51]** | ESI-MS | proteins | other |

Other algorithms not mentioned in *Table 1* are as follows: probabilistic classifier with dynamic programming [31], a statistical approach based on a non-negative sparse regression scheme [32], LASSO method [33], *Pepex* based on quadratic programming [34], approximation of isotopic envelopes using a Poisson distribution [35], *Isotopica* based on the Fast Fourier Transform [37], algorithm based on computing isotopic patterns from the elemental composition of the product ions' actual amino acids [38]. The algorithms have some limitations: requiring the knowledge of the parameters for optimization [34] [39], those based on comparing the theoretical and experimental isotopic envelope have difficulties with detecting overlapping isotopic envelopes because of relying solely on intensities [39], *Decon2LS* and probabilistic classifier do not take into consideration the overlapping isotopic envelopes [31] [42]. In comparison, the statistical approach [32] decreases the number of false positives and false negatives, as it selects only the simplest model that has a smaller number of isotopic envelopes [32] [39].

In *Table 1*, common deisotoping algorithms are listed, and one can notice that only a few methods are dedicated to MALDI-ToF-MS data. It is worth mentioning that MALDI-MSI datasets require efficient methods to be applied for data analysis, as they can have over 40 GB.

*Data characteristics.* Peptides data (mass spectra acquired in MALDI-ToF MSI experiments) was collected by Maria Skłodowska-Curie National Research Institute of Oncology in Gliwice (Poland), which came from a patient with head and neck cancer (HNC) - oral cavity squamous cell carcinoma, cancer stage T4N2M0. Four datasets are from a fresh-frozen tissue sample (FF, published in [9]). The other four datasets are from Fomalin-Fixed Parafin-Embedded (FFPE), published in [10]. MSI experiment workflow, tissue preparation and preservation is presented in *Figure 1*.

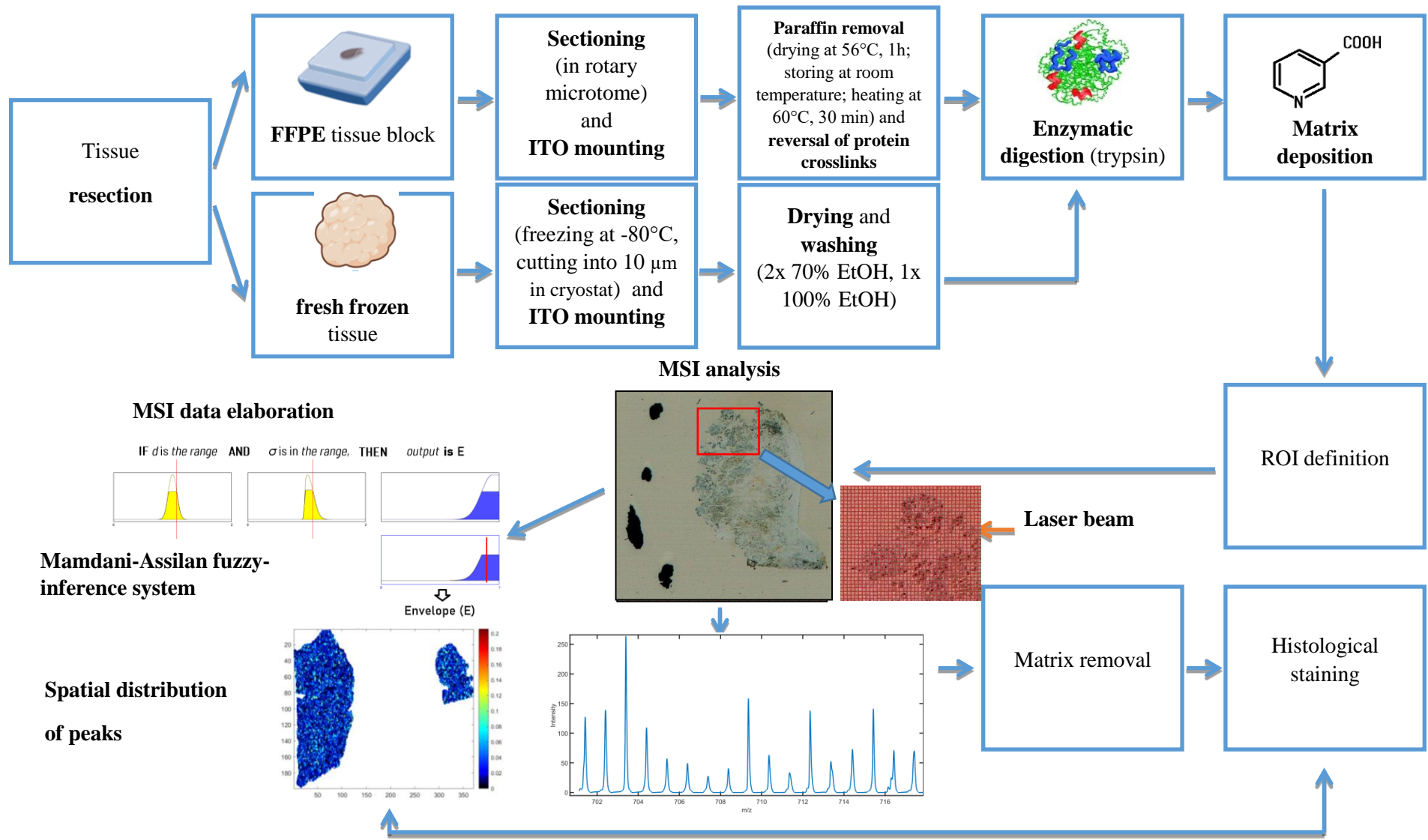

*Figure 1.* MSI experiment workflow for FF and FFPE tissue block based on [54].

After that, both HNC-FF and HNC-FFPE datasets underwent the pre-processing and feature extraction pipeline comprised of the following steps: unification of mass channels, noise reduction, adaptive baseline correction, outlying spectra identification, spectra alignment, normalisation to the mean TIC [41][42][11], average spectrum modelling, peak detection using the Gaussian Mixture Modelling (GMM), removal of GMM components characterised by high variance and/or low amplitude, modelling the right-skewed spectrum peaks with the use of GMM components, merging with the left-neighbouring major component, calculating peak intensity (abundance) by the pairwise convolution of GMM components and individual spectra [11-14]. The Gaussian component is defined by its location, shape parameter, and the area under the curve. Thus, in further analysis, calculations are performed using obtained Gaussian components and their parameters, but the term *peak,* widely used when analysing the mass spectra, will be used interchangeably.

HNC-FF [9]

The mass spectrometer (ultrafleXtreme MALDI-ToF, Bruker Daltonik, Bremen, Germany) worked in the positive reflectron mode. The masses were acquired in the 800 – 4000 m/z with a raster width of 100 μm. After the mass spectrometry experiment, the matrix was removed from the slides, and the tissue specimens were H&E stained [9]. The experiment was described and published in [9]. The original dataset comprised 45 738 raw mass spectra with 109 568 mass channels (m/z). After undergoing the pre-processing and feature extraction steps, 3 714 components remain.

HNC-FFPE [10]

After the on-tissue digestion step and before coating with the matrix, the tissue section was placed in a humid chamber in a solution: 100 mM NH4HCO3M, 5% MeOH at 37 °C, 18h). The mass spectrometer (ultrafleXtreme MALDI-ToF, Bruker Daltonik, Bremen, Germany) worked in the positive reflectron mode, the masses were acquired in the range of 700 – 3000 m/z with a raster width of 100 μm [10]. The datasets comprised 22 389, 22 267, 21 395, and 31 654 raw mass spectra with 200 704 mass channels, respectively. After pre-processing and feature extraction, the number of components (tryptic peptide species) decreased to 1 776, 1 766, 1 697, and 2 510, respectively. The experiment was described and published in [10].

Potential isotopic envelopes annotation

Potential isotopic envelopes were annotated by an experienced mass spectrometrist.

*Idea detailed explanation.* In the sample, many natural isotopes can be observed since the vast majority of elements have different isotopic forms in nature [15]. The isotopic envelope consists of signals originating from molecules of the same analyte but containing isotopes of elements differing in weight (e.g., $^{12}$C vs. $^{13}$C).

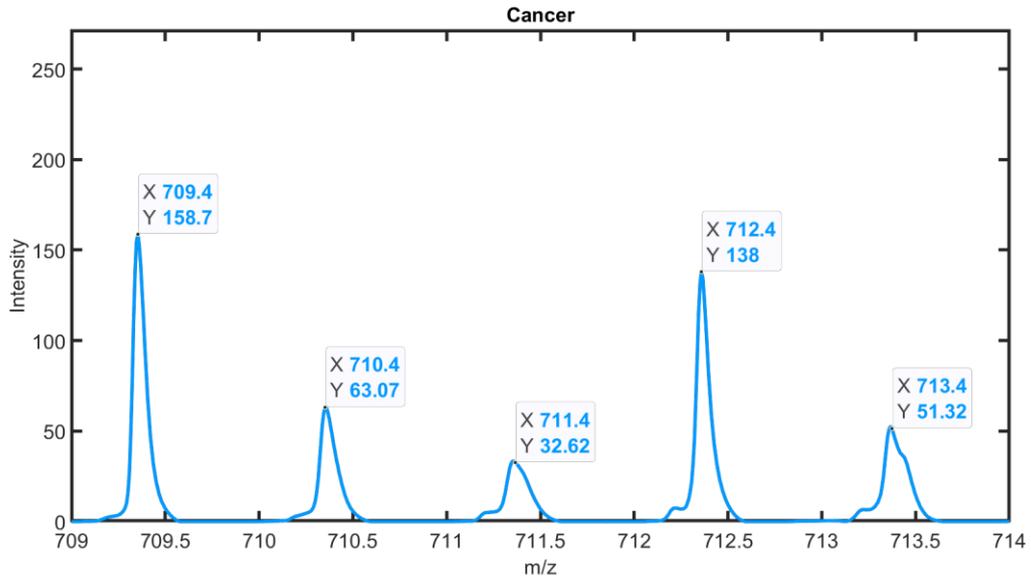

***Figure 2.*** Peptides mass spectrum with an exemplary isotopic envelope.

Two types of isotopic envelopes can be distinguished:

a) non-overlapping (*Figure 3*)

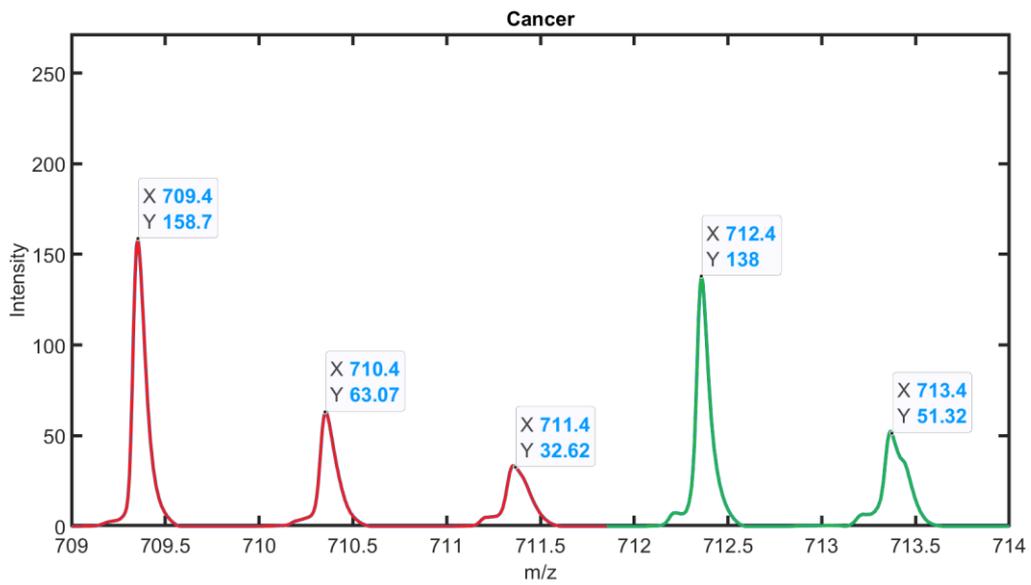

**Figure 3.** Example of non-overlapping isotopic envelopes **(red** – first isotopic envelope, **green** – the second one).

b) Overlapping *(Figure 4)*

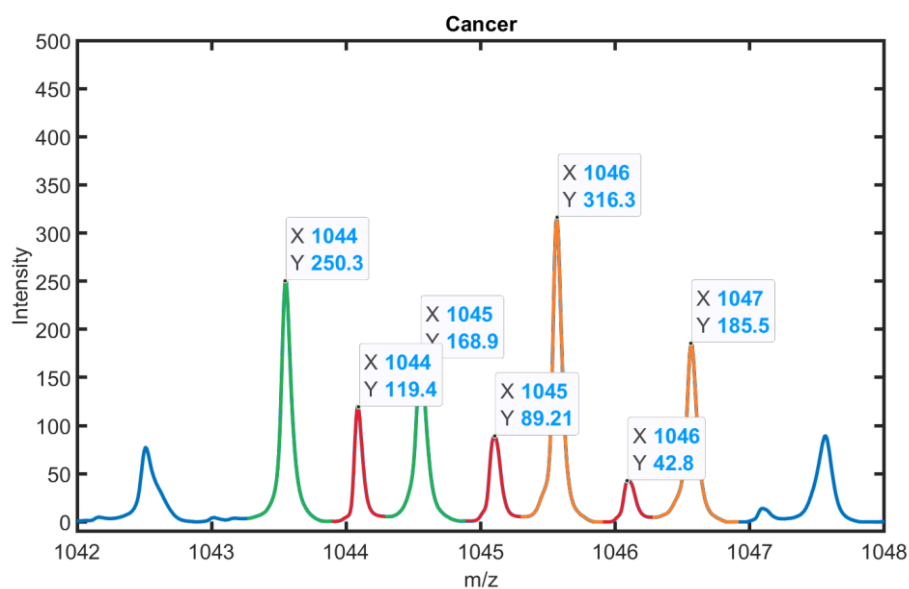

**Figure 4.** Overlapping isotopic envelopes (**green** – first isotopic envelope, **red** – second, **orange** – the third one).

In order to identify peaks that are the members of isotopic envelopes, the following algorithm has been employed (*Figure 5*):

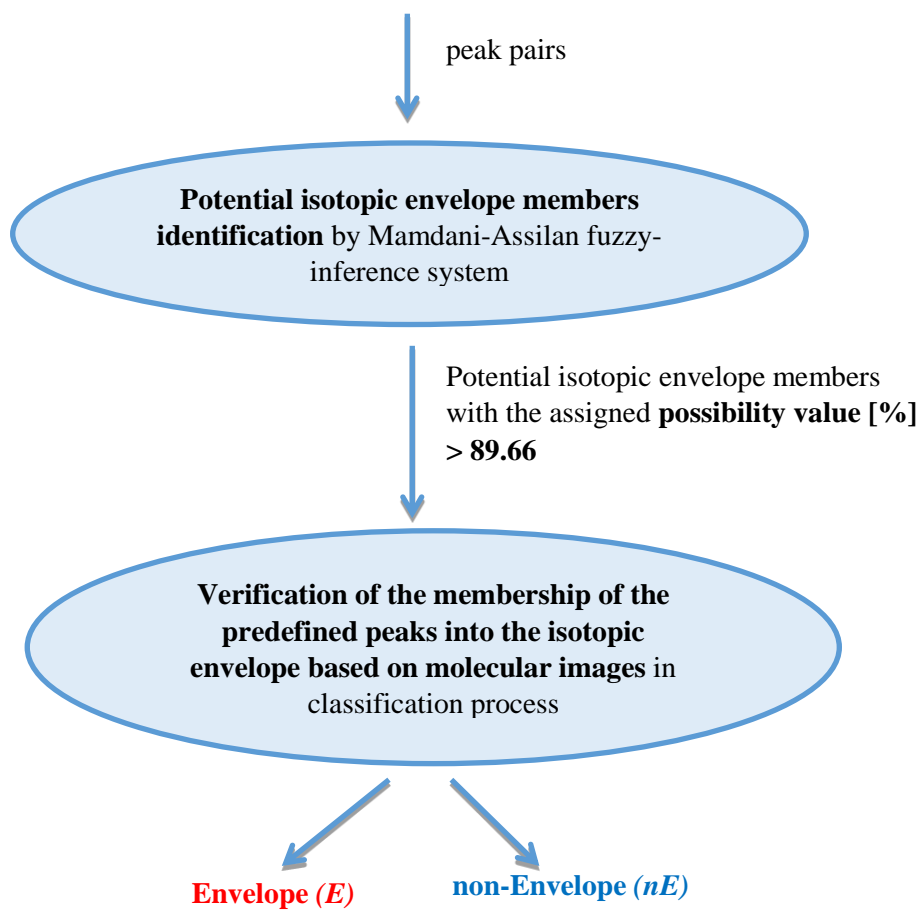

**Figure 5.** Algorithm workflow.

<u>First step:</u> preselection of potential isotopic envelope members identification by Mamdani-Assilan fuzzy-inference system

The idea of such a system has already been published in [16], but it has significantly changed and is presented in this work. This system was built on the following assumptions:

1) Distance between means of two adjacent model components is approximately equal to one [16]

Typically MALDI data consist of single charged ions on the mass spectrum, therefore [16]:

$$\frac{1.003}{z} = \frac{1.003}{1} = 1.003 \text{ [Da]} \tag{1}$$

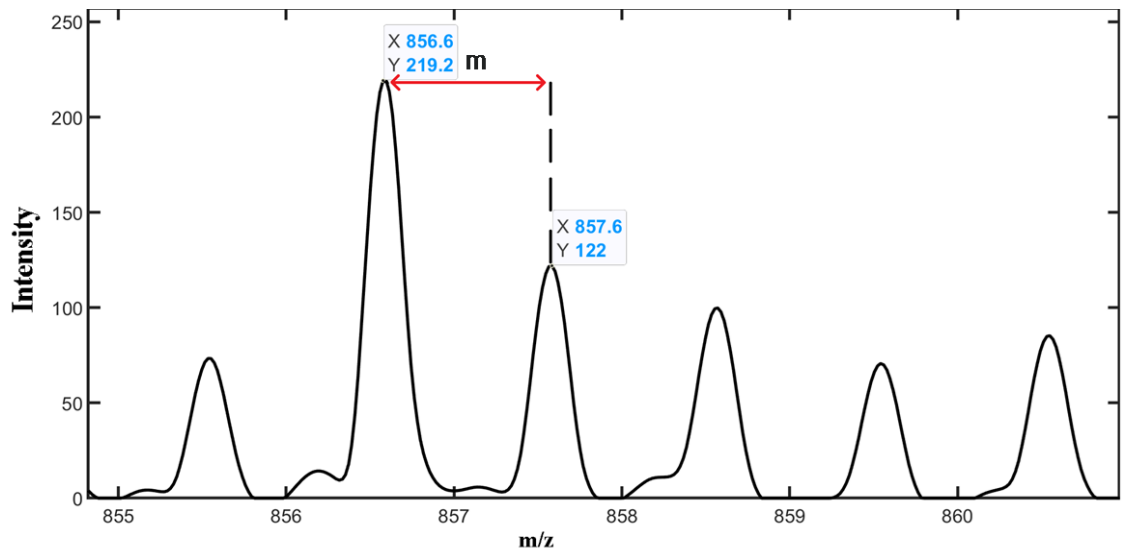

**Figure 6.** Distance *(m)* between means of two adjacent model components.

2) Ratio of estimated variances of adjacent model components is approximately equal to one: [16]

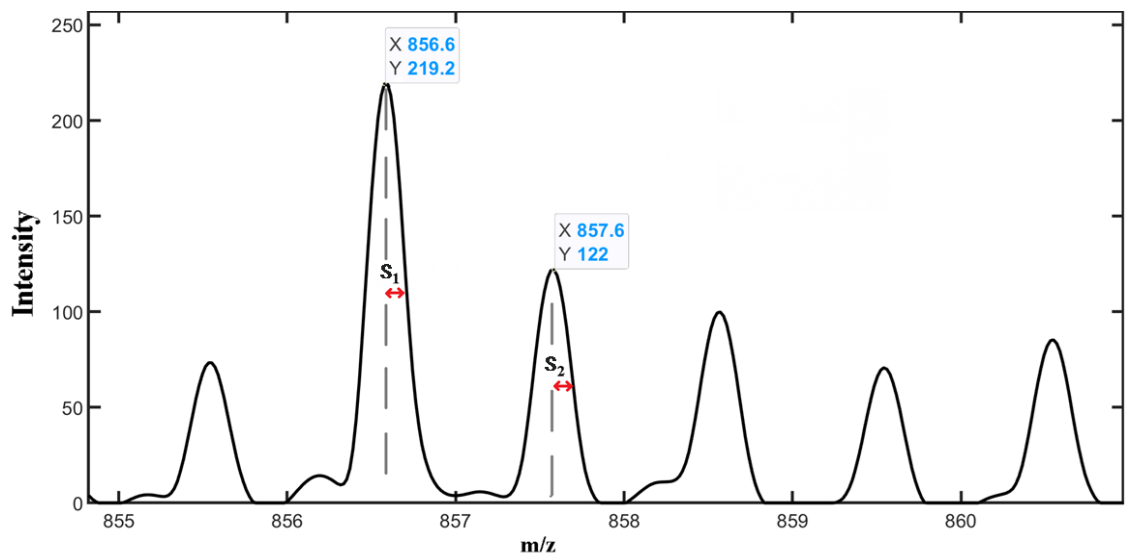

**Figure 7.** Ratio of variances *(s)* of the adjacent model components.

The fuzzy-inference system has the following structure (*Figure 8*): two inputs (*m* and *s*), one output, membership functions type: Gaussian.

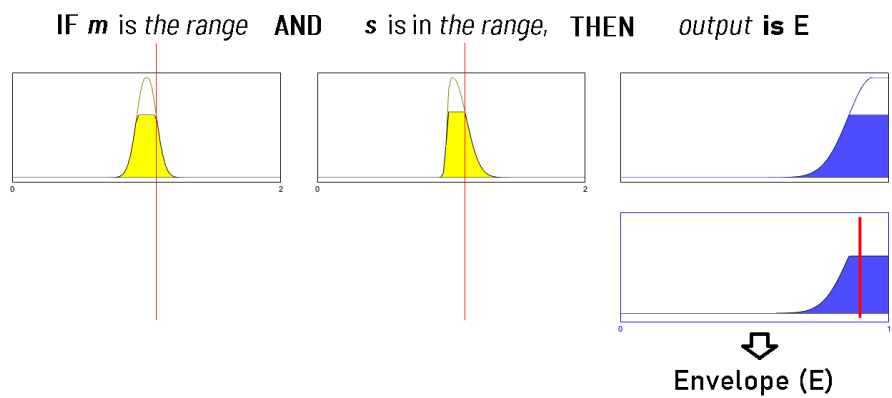

**Figure 8.** Implication and aggregation in created fuzzy-inference system.

Every pair of peaks is assigned the so-called *possibility value* [%], which indicates whether the given pair of peaks is included in an envelope *(E)* or not included *(nE)* (*Table 2*).

**Table 2.** Exemplary results.

| m/z$_1$ | m/z$_2$ | Possibility of isotopic envelope membership [%] |
|---|---|---|
| 805.6 | 809.7 | 46.0 **(non-Envelope)** |
| 808.7 | 809.7 | 74.7 **(non-Envelope)** |
| 810.7 | 811.7 | 98.1 **(Envelope)** |
| 810.8 | 897.6 | 15.3 **(non-Envelope)** |
| 812.7 | 813.7 | 98.7 **(Envelope)** |
| 812.7 | 897.6 | 25.1 **(non-Envelope)** |
| 843.7 | 844.7 | 99.0 **(Envelope)** |

In order to define the threshold value over which the possibility of isotopic envelope membership value indicates that this pair of peaks is included in an isotopic envelope, the Gaussian Mixture decomposition was applied [18-19]; the number of gradients was calculated using BIC [20] – it was decided to choose five components, as after that value the slope is not changing significantly (*Figure 9*).

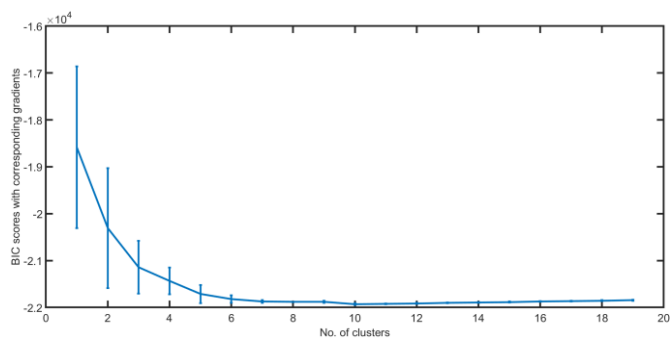

**Figure 9.** BIC scores with corresponding gradients vs. number of clusters.

According to *Figure 10*, 0.8966 is the threshold for distinguishing isotopic envelopes peaks from non-envelope ones.

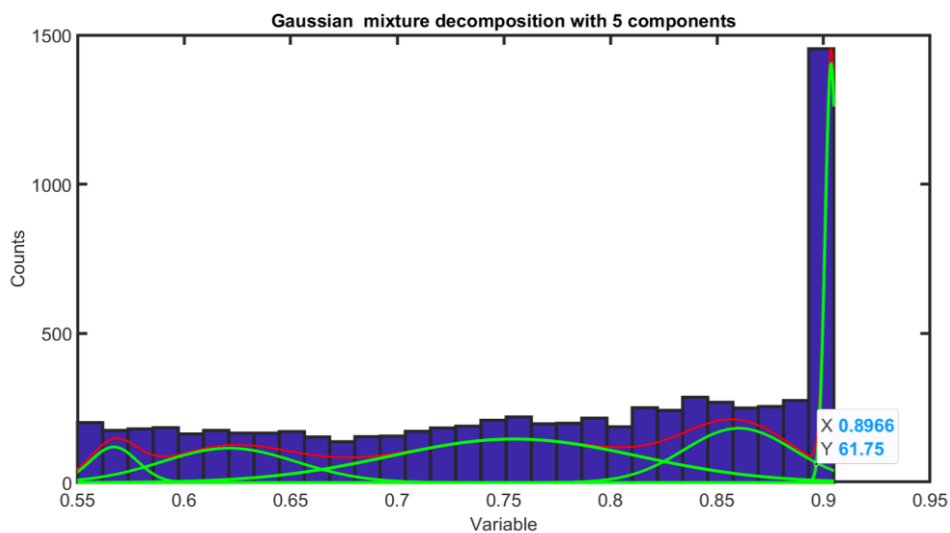

**Figure 10.** GMM decomposition with 5 components.

The main goal of this step was to reduce the number of peak pairs that should undergo the isotopic envelope member selection process.

**Table 3.** Results after employing fuzzy-inference system for potential isotopic envelope peaks identification.

| Dataset Name | No. of input peak pairs | No. of potential isotopic envelope member peak pairs | %of peak pairs reduction ↓ |
|---|---|---|---|
| **HNC-FFPE Dataset 1** | 97 910 | 1 916 | 98.04 |
| **HNC-FFPE Dataset 2** | 55 030 | 1 457 | 97.35 |
| **HNC-FFPE Dataset 3** | 68 920 | 1 584 | 97.70 |
| **HNC-FFPE Dataset 4** | 97 840 | 1 945 | 98.01 |
| **HNC-FF Dataset 1** | 47 750 | 1 662 | 96.52 |
| **HNC-FF Dataset 2** | 46 350 | 1 610 | 96.53 |
| **HNC-FF Dataset 3** | 54 030 | 1 624 | 96.99 |
| **HNC-FF Dataset 4** | 55 070 | 1 603 | 97.09 |

As a result, the potential isotopic envelopes' member peaks are defined (*Figure 11*) (*Table 3*). It can be noticed that the number of the peak pairs that were preselected as being isotopic envelope members has been significantly diminished.

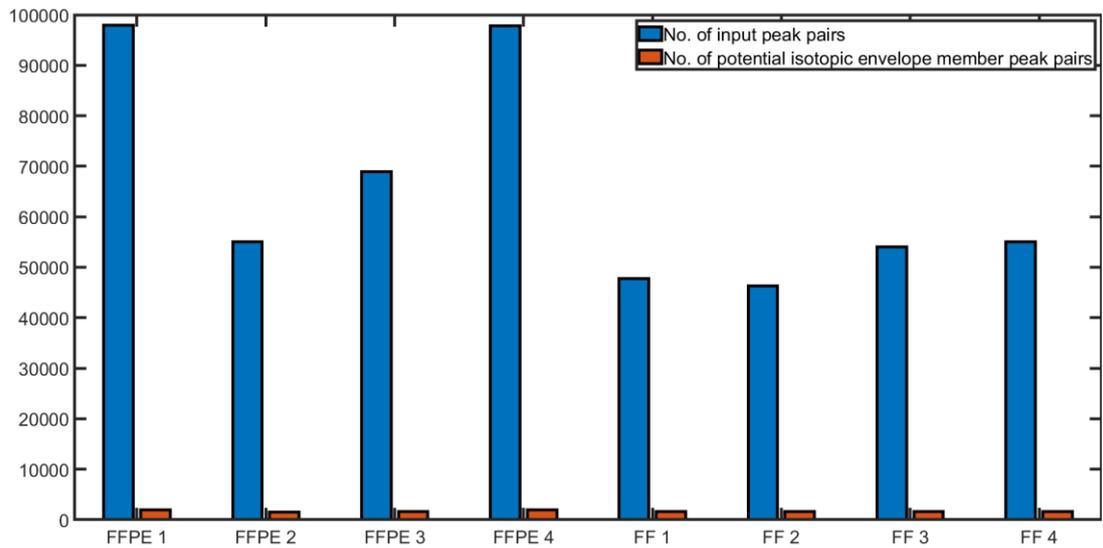

**Figure 11.** Comparison of number of input peak pairs before applying fuzzy-inference system vs. after.

<u>Second step</u>: verification of the membership of the predefined peaks into the isotopic envelope

This part is based on mapping the spatial distribution of an analyte. Every peak from the mass spectrum with a given m/z value is visualized as a map of intensity (spatial map of a molecular distribution) (*Figure 12*), which reflects the peak intensities registered for every m/z across the whole tissue section. This idea has already been announced in [17].

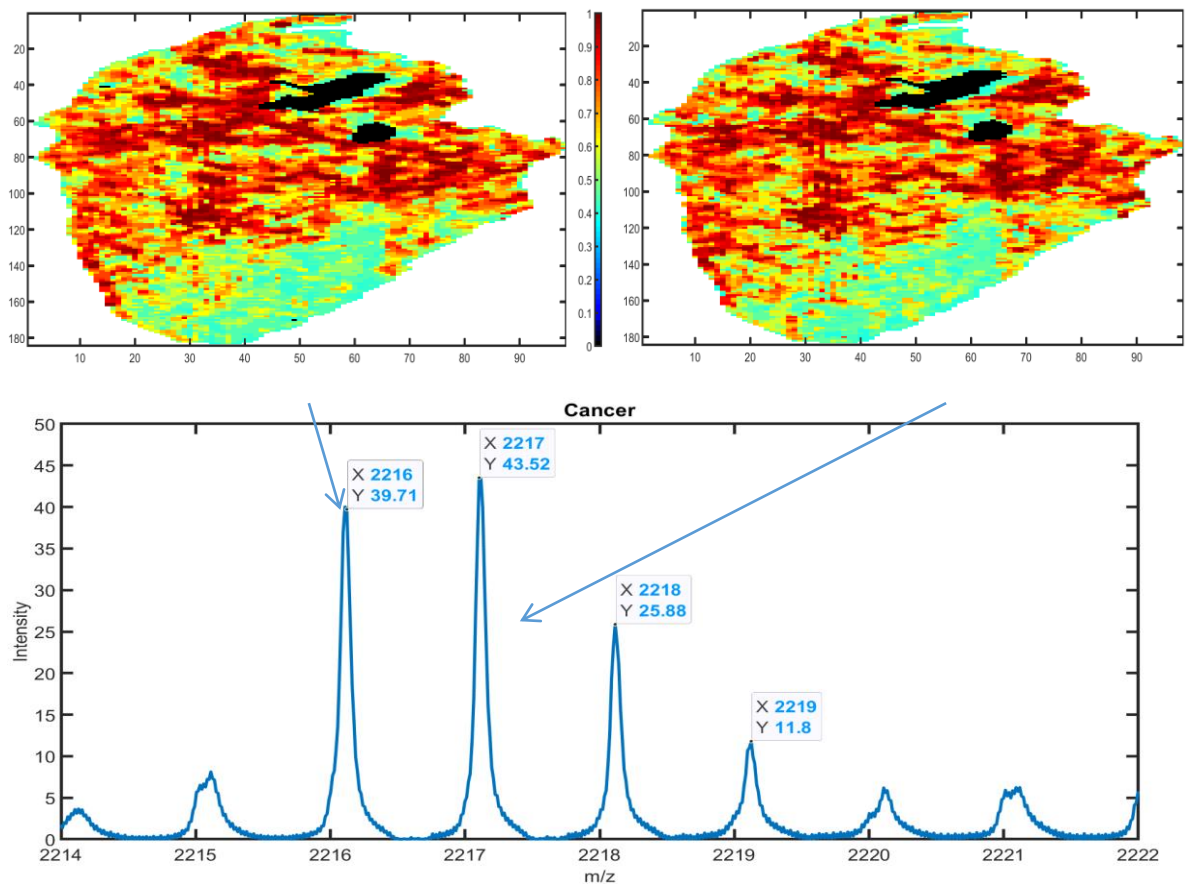

**Figure 12.** Visualisation of the peaks with given m/z values as the spatial maps of molecular distribution (maps of intensities).

Every constructed map was handled as a digital image, and the following steps were applied: histogram enhancement, filtering, and normalisation. Then, the differential images were constructed by subtracting images of separate peaks |IMG 1 – IMG 2| in the following way *(Figure 15, Figure 18)* [17]:

**Envelope**

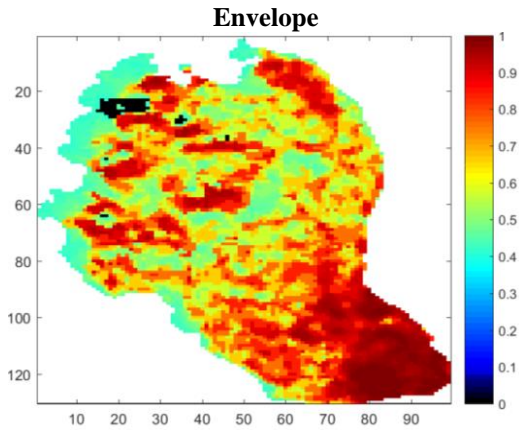

*Figure 13. Image A.*

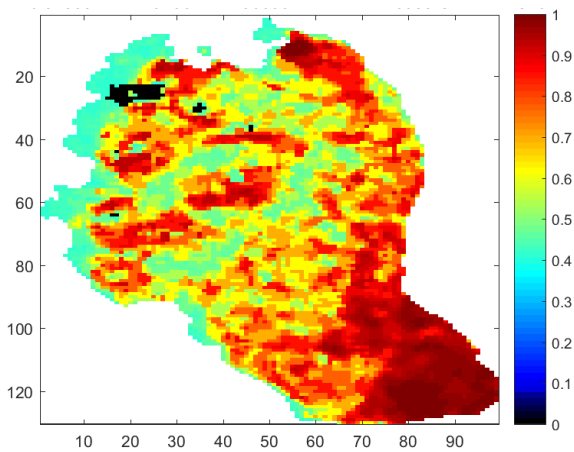

*Figure 14. Image B.*

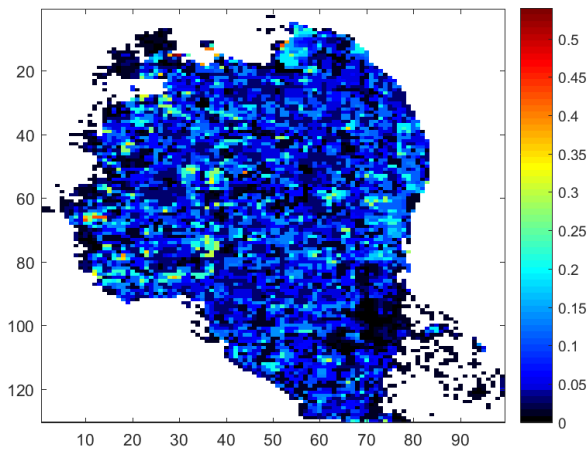

*Figure 15. Differential image A – B.*

**Non-envelope**

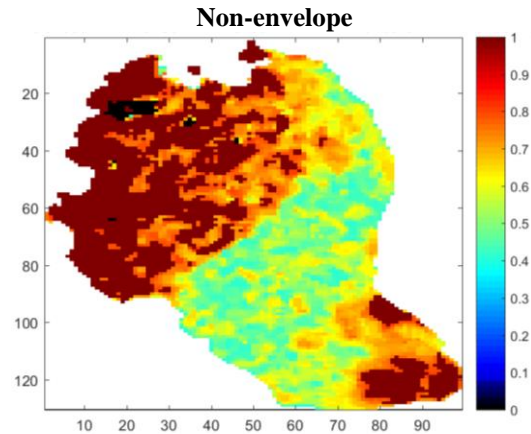

*Figure 16. Image C.*

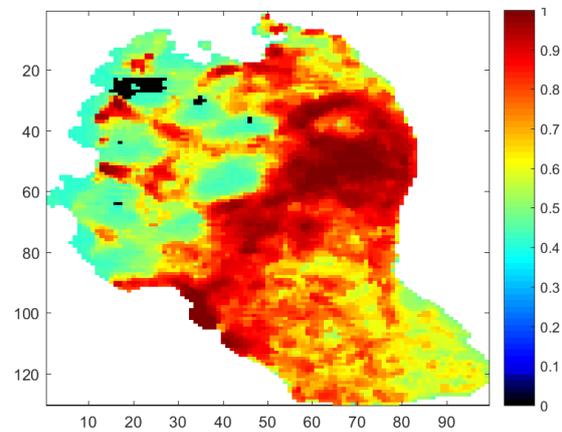

*Figure 17. Image D.*

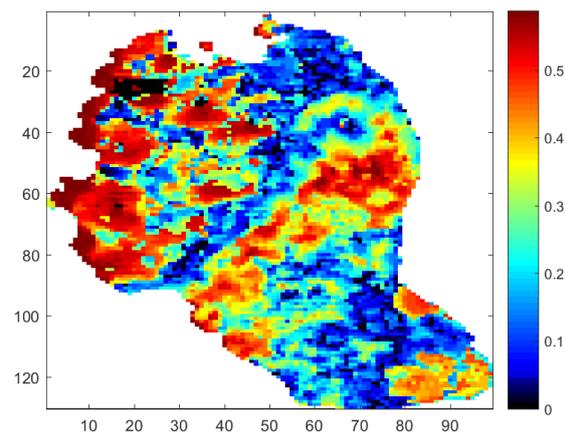

*Figure 18. Differential image C – D.*

Such a visualisation shows the differences between the spatial distribution of $E$ and $nE$ peaks. We claim that if two peaks are included in the isotopic envelope, their intensity structure and structural image properties are similar. Therefore, after subtracting their spatial maps of intensities *(Figure 13, Figure 14, Figure 15)*, no structure should be visible, as a resulting image is characterised by uniform intensity distribution (or the intensities are at a similar level) [17, 21]. Otherwise, the peaks are not included in the isotopic envelope since the similar spatial distribution is not similar. Thus, structurality in a differential image is clearly visible *(Figure 16, Figure 17, Figure 18)* [17].

In order to assign peaks to the envelope (*E*) groups or non-Envelope (*nE*), several metrics based on the spatial map of intensities were applied based on texture metrics. Some of them are based on the *Gray-Level Co-Occurrence matrix (GLCM) (Figure 20),* and some of them are based on differential images, unlike others *(Figure 19).*

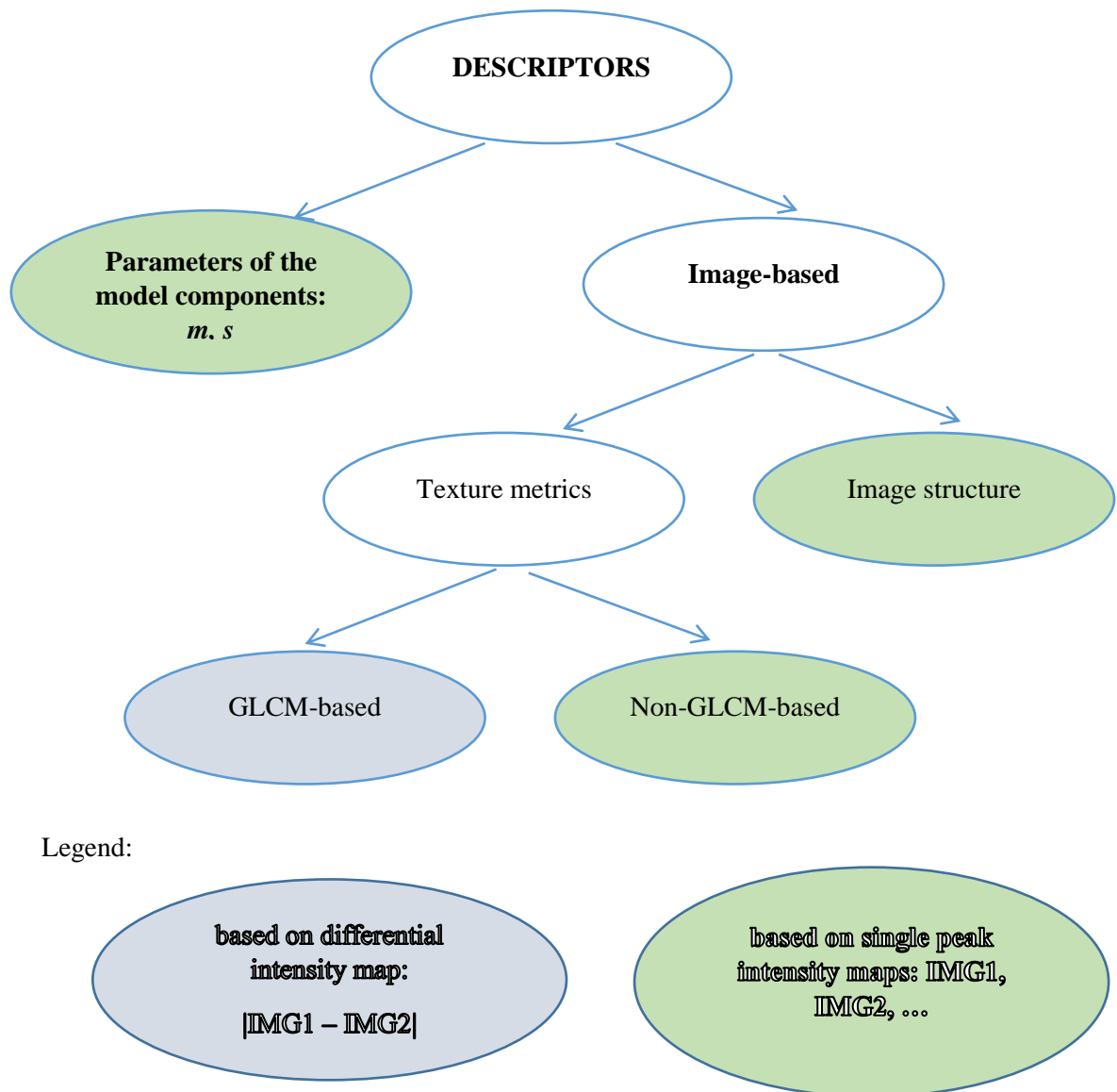

**Figure 19.** Descriptors divided into groups and distinguished based on the calculation method: differential image-based and based on separate peaks images.

| Contrast focused | Order type | Statistical | Non-GLCM-based |
|---|---|---|---|
| • contrast<br>• homogeneity<br>• M1-based | • entropy<br>• energy | • mean<br>• standard deviation<br>• variance<br>• correlation<br>• moment<br>• median<br>• interquartile range (IQR)<br>• coefficient of variation (cV) | • autocorrelation |

**Figure 20.** Texture metrics groups.

From every group presented in *Figure 20* each feature was analysed in the similar way *(Figure 21, Figure 22)*. Results are presented in *Supplementary Material*.

Contrast is a measure of the texture grooves' depth [22-23].

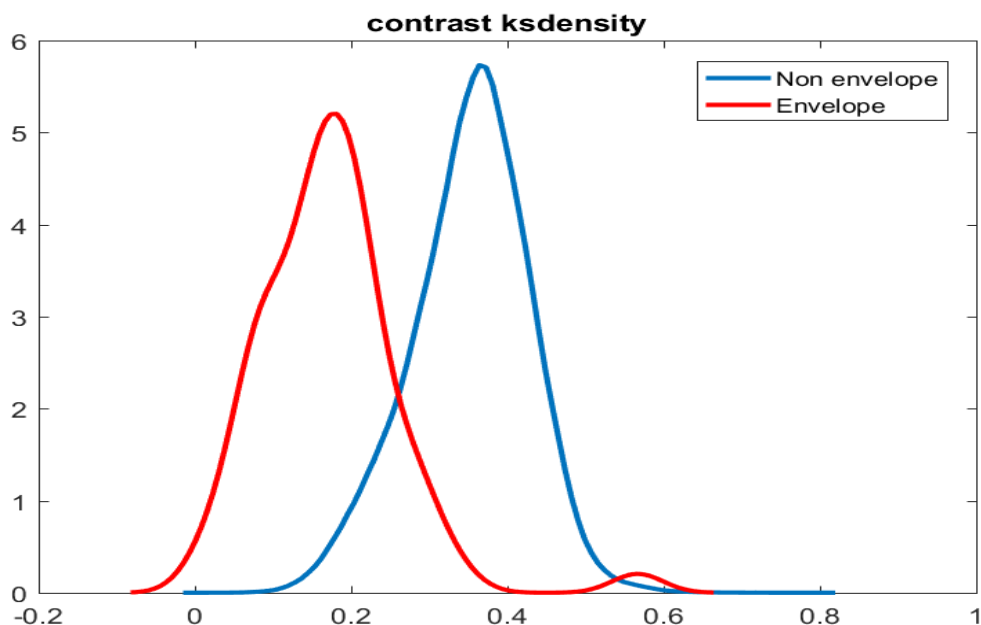

**Figure 21.** Contrast probability density estimate for the training data (envelope and non-envelope peaks).

It can be noticed that over 95% of envelopes are in the range (0; 0.3), whereas non-envelopes in the same range are about 15%.

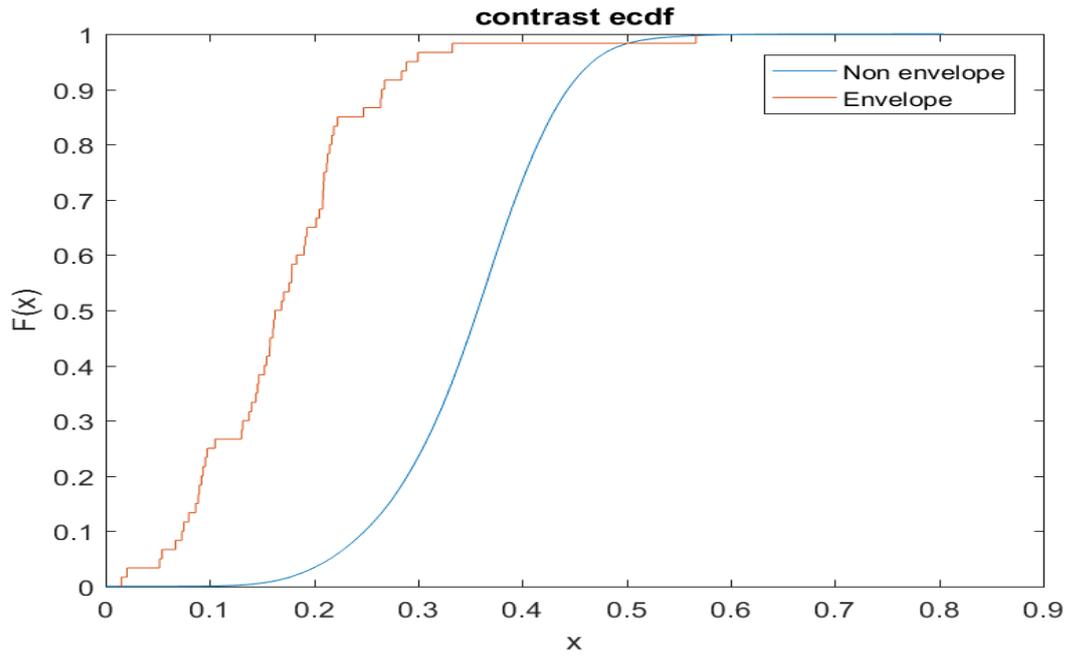

**Figure 22.** The empirical cumulative distribution function of contrast metric for the envelope and non-envelope peaks.

After analysis of all the features on their reference to the envelope identification, the following summary has been proposed (*Figure 23*):

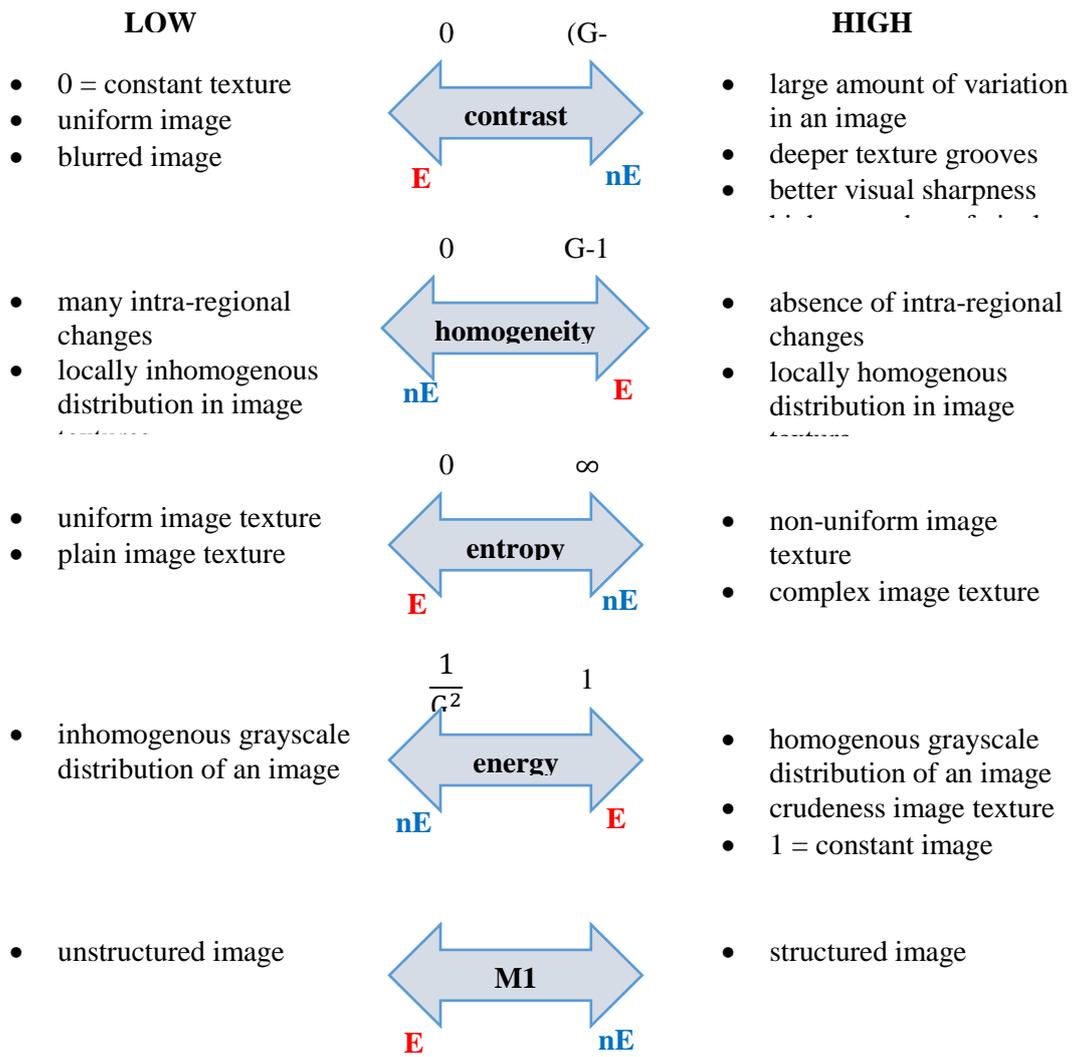

**Figure 23.** Image texture metrics from contrast- and order-type groups interpretation in relation to isotopic envelopes membership. Interpretation based on [24-27].

In order to assess the correlation between image texture metrics, Spearman's rank correlation has been calculated (*Figure 24*). Since many presented features are calculated similarly, they are strongly correlated [25]. Contrast group measures are strongly correlated to variance, whereas between contrast and homogeneity, a negative correlation value is expected ($r = -0.8623$) [25] (contrast $r = 0.7487$, homogeneity $r = -0.6327$ and M1 $r = 0.7068$). Entropy seems more independent than other features ($r < 0.5194$ for every feature). Consequently, those features can be combined, and at least one of each group should be chosen for further classification [25].

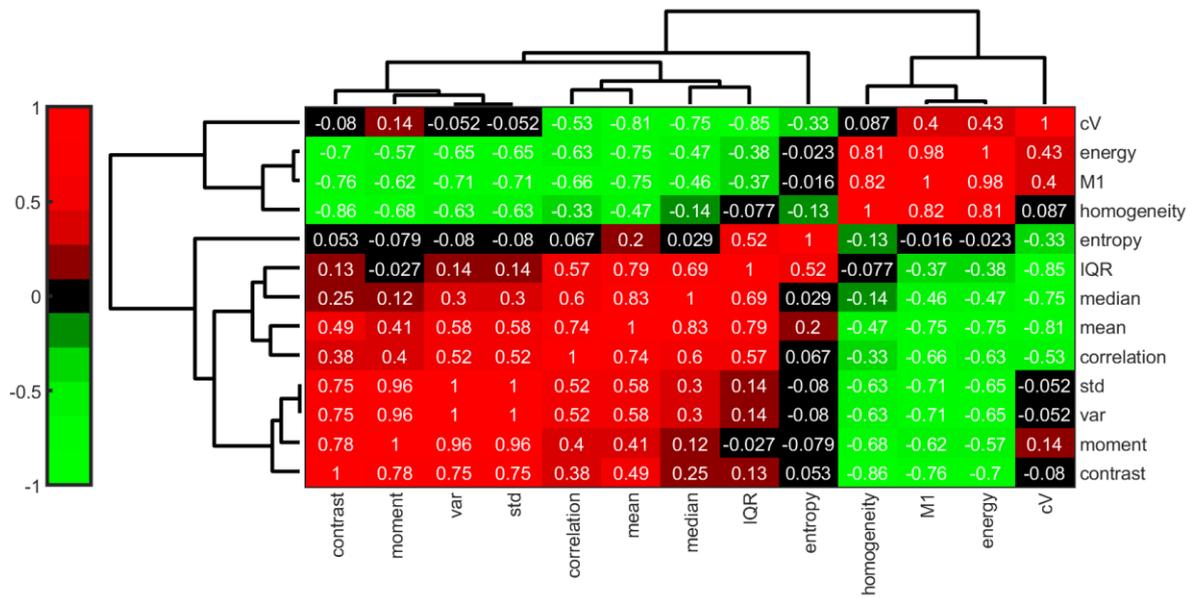

**Figure 24.** Clustergram of Spearman's rank correlation coefficient of the image texture metrics.

After choosing the descriptors, feature selection method based on *forward propagation* [28] was applied, resulting in choosing 8 out of 17 descriptors for further analysis (*Figure 25*): *m, s, correlation, entropy, median, contrast, homogeneity, moment*.

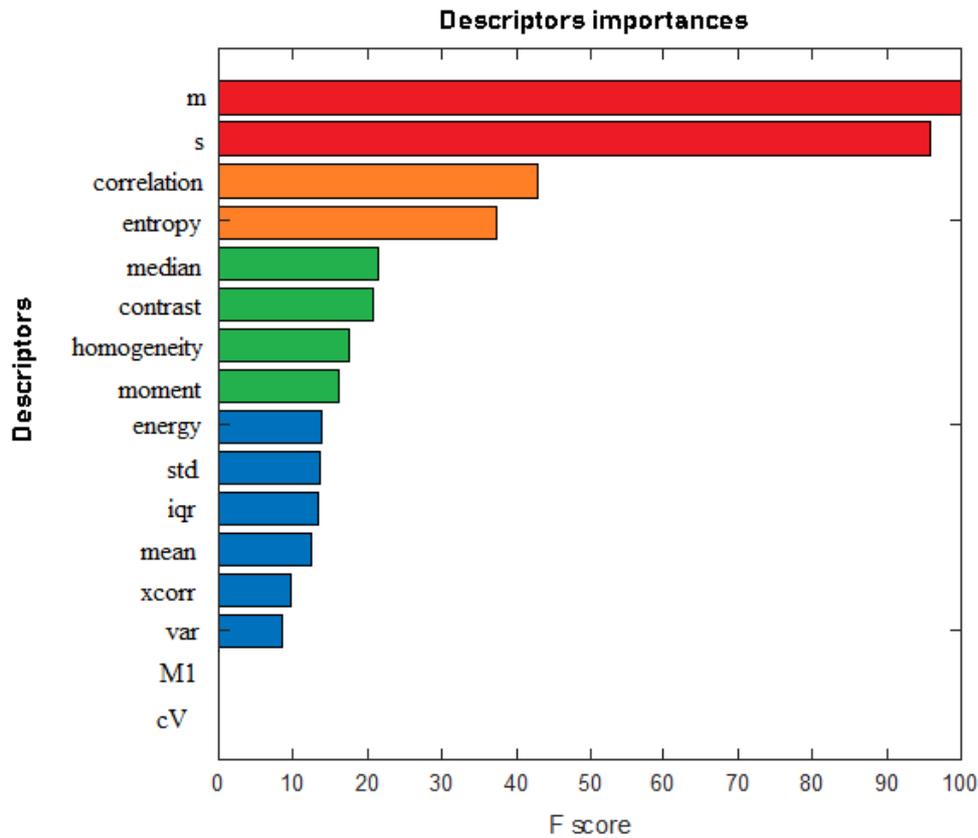

**Figure 25.** Clustergram of Spearman's rank correlation coefficient of the image texture metrics.

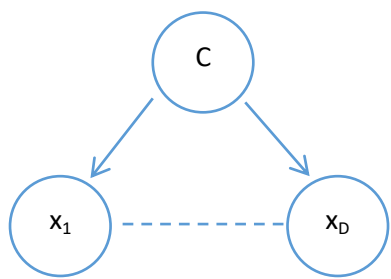

**Figure 26.** A graphical representation of the naive Bayes model for classification [29].

In order to classify peaks to the one of the class: *E* and *nE*, the supervised learning approach – Naïve Bayes classifier with Epanechnikov kernel function was employed (*Figure 26*).

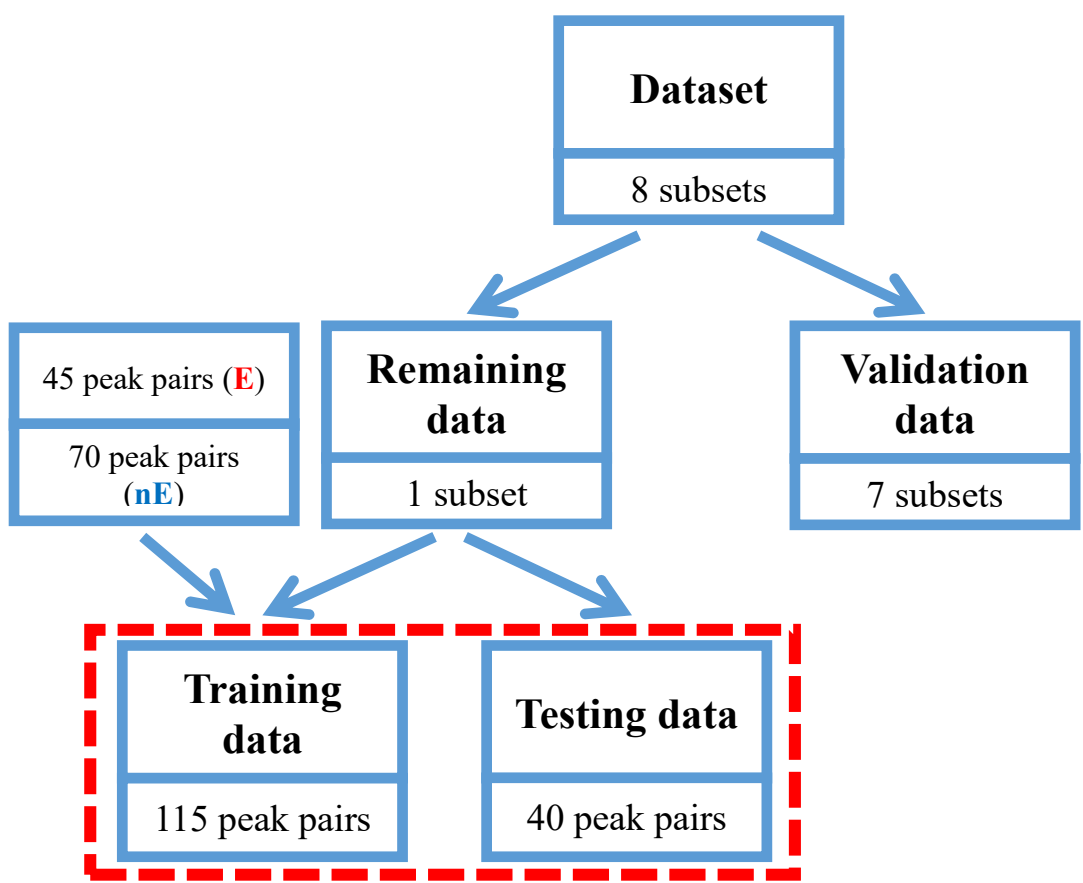

**Figure 27.** Dataset division scheme.

The model was built by dividing the data into training and testing (*Figure 27*). Validation data was not included in the process of model creation. Training data was used to fit a model and was created in the following way: the matrix of differential images was created. Each differential image was based on subtracting two peaks from each other. Once an expert had annotated the peaks, we could, at this stage, assign exactly one label to each cell in the matrix, whether or not the differential image came from peaks that were members of an isotopic envelope (**E**) or not (**nE**). To validate the model, cross-validation, with the division of the entire dataset into five subsets, was performed. Then, the model was trained 100 times on each subset. The performance measure (accuracy) of overall training subsets was calculated.

# RESULTS

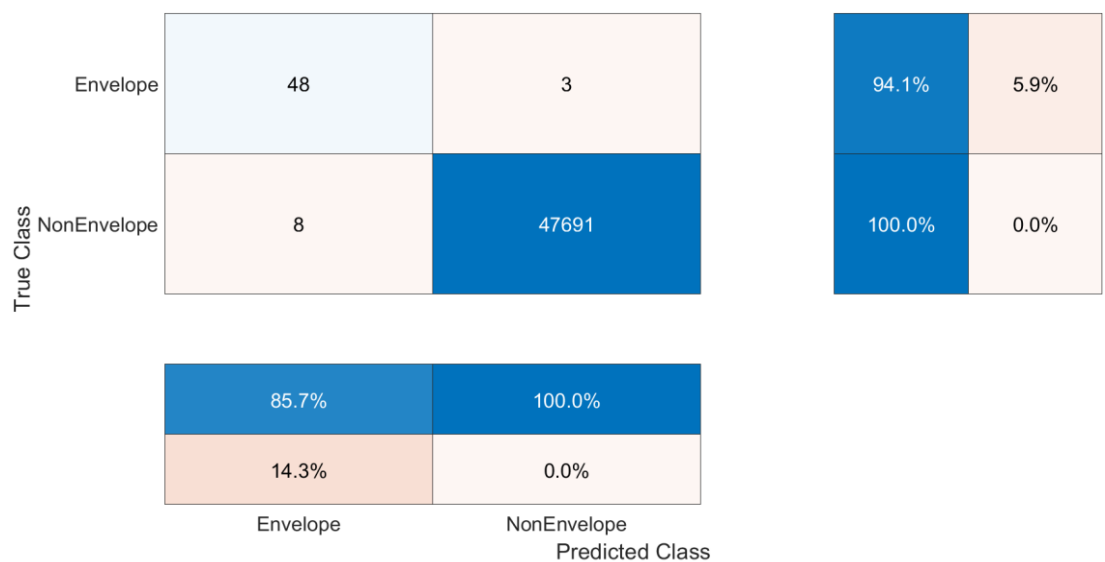

**Figure 28.** *Confusion matrix for HNC-FF Daset 1.*

As can be observed in *Figure 28* and *Figure 29*, the classifier is characterized by significant agreement between the expert annotation of envelope peaks and the classifier (48 and 242 identified envelopes, respectively). It is worth mentioning that the classifier annotated more isotopic envelopes peaks than the expert. Consequently, the information on possible isotopic envelopes will be recovered, as the expert is not sure if a peak is a member of an isotopic envelope in every case due to the signal density.

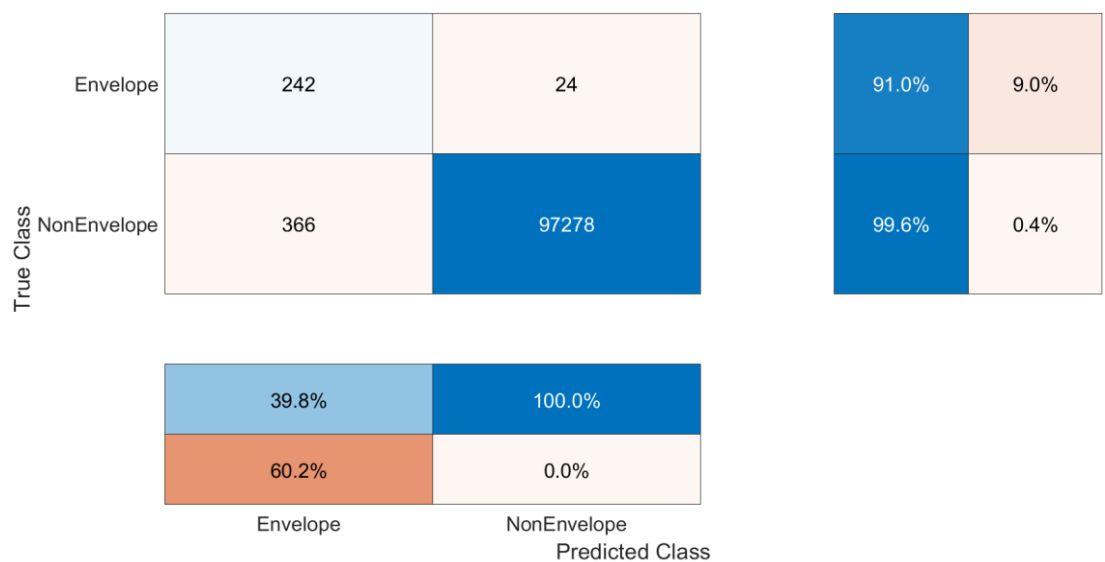

**Figure 29.** Confusion matrix for *HNC-FFPE Daset 1*.

According to *Table 4*, recall and precision values ranging from 88.12% - 94.12% indicate that peaks included in isotopic envelopes are classified accurately. A specificity value over 99% means that peaks not included in isotopic envelopes were classified with such high accuracy. Regardless of the number of elements included in E and nE classes, the model quality can be assessed by Matthews Correlation Coefficient, which ranges from 80.14% to 89.81%. Moreover, balanced accuracy is also significant: 94.03% - 97.05%. Another indicator of an agreement between predicted values and those assessed by the expert is Fowlkes-Mallows Index, which indicates a significant similarity (over 80%).

Table 4. Statistic metrics.

|  | HNC-FF Dataset 1 | HNC-FF Dataset 2 | HNC-FF Dataset 3 | HNC-FF Dataset 4 |
| --- | --- | --- | --- | --- |
| **TP** | 48 | 45 | 89 | 128 |
| **TN** | 47691 | 46289 | 53896 | 54900 |
| **FN** | 3 | 6 | 12 | 9 |
| **FP** | 8 | 10 | 33 | 33 |
| **Specificity [%]** | 99.98 | 99.98 | 99.94 | 99.94 |
| **Precision** | 85.71 | 81.82 | 72.95 | 79.50 |
| **Recall** | 94.12 | 88.24 | 88.12 | 93.43 |
| **Balanced Accuracy** | 97.05 | 94.12 | 94.03 | 96.69 |
| **Critical Success Index** | 81.36 | 73.77 | 66.42 | 75.29 |
| **Matthews Correlation Coefficient** | 89.81 | 84.95 | 80.14 | 86.15 |
| **Prevalence Threshold** | 1.32 | 1.54 | 2.57 | 2.47 |
| **Fowlkes-Mallows Index** | 89.82 | 84.97 | 80.18 | 86.19 |

Analysing *Figure 30* of three overlapping isotopic envelopes, it can be observed what was proven by the values in confusion matrix that our algorithm identified more isotopic envelopes peaks than the expert.

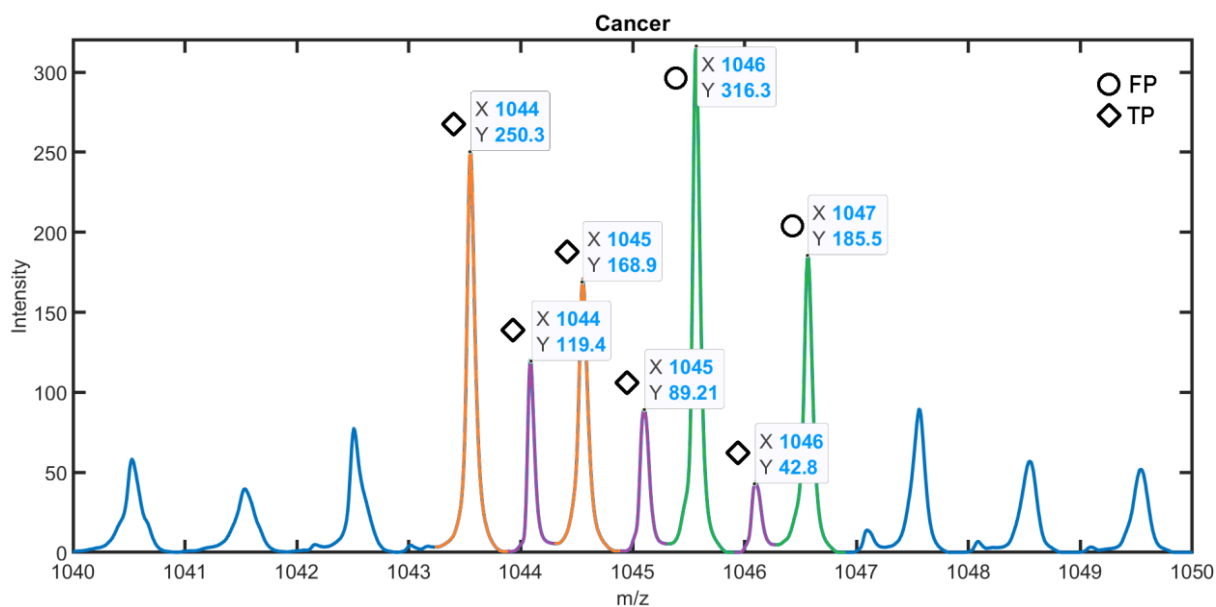

**Figure 30.** Exemplary overlapping isotopic envelopes (**orange** – first isotopic envelope, **violet** – second isotopic envelope, **green** – third isotopic envelope). The reference for FP (False Positive) and TP (True Positive) was the expert annotation.

The assumption based on no visible structurality presence in the spatial distribution maps of peaks included in one isotopic envelope has also been proved, according to the below images *(Figure 31, Figure 32, Figure 33, Figure 34)*:

**Orange envelope:**

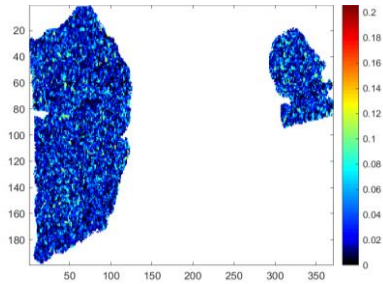

**Figure 31.** Differential image: | m/z$_1$ – m/z$_2$ | = |1044 – 1045|.

**Violet envelope:**

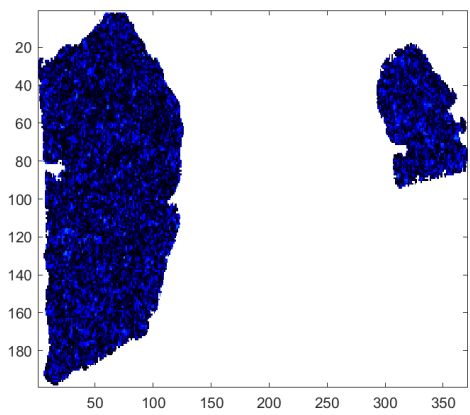 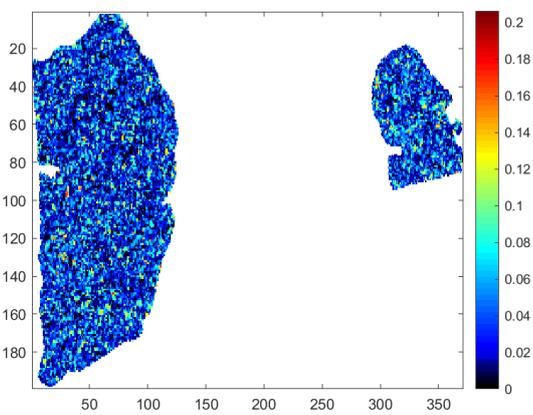

**Figure 32.** Differential image: |1044 – 1045|.     **Figure 33.** Differential image: |1045 – 1046|.

**Green envelope:**

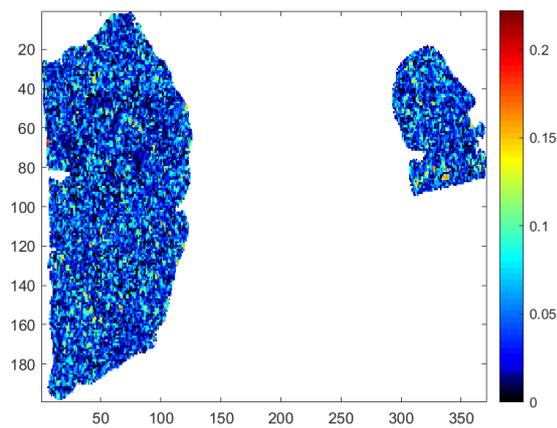

**Figure 34.** Differential image: |1046 – 1047|.

**Table 5.** Number of detected isotopic envelopes with a given length in *HNC-FFPE Dataset 1*.

| No. of peaks included in an isotopic envelope | No. of detected isotopic envelopes |
|---|---|
| 2 | 295 |
| 3 | 64 |
| 4 | 38 |
| 5 | 11 |
| 6 | 2 |
| 7 | 1 |
| 11 | 1 |

In *Table 5*, it can be noticed that the vast majority of isotopic envelopes consist of two members peaks.

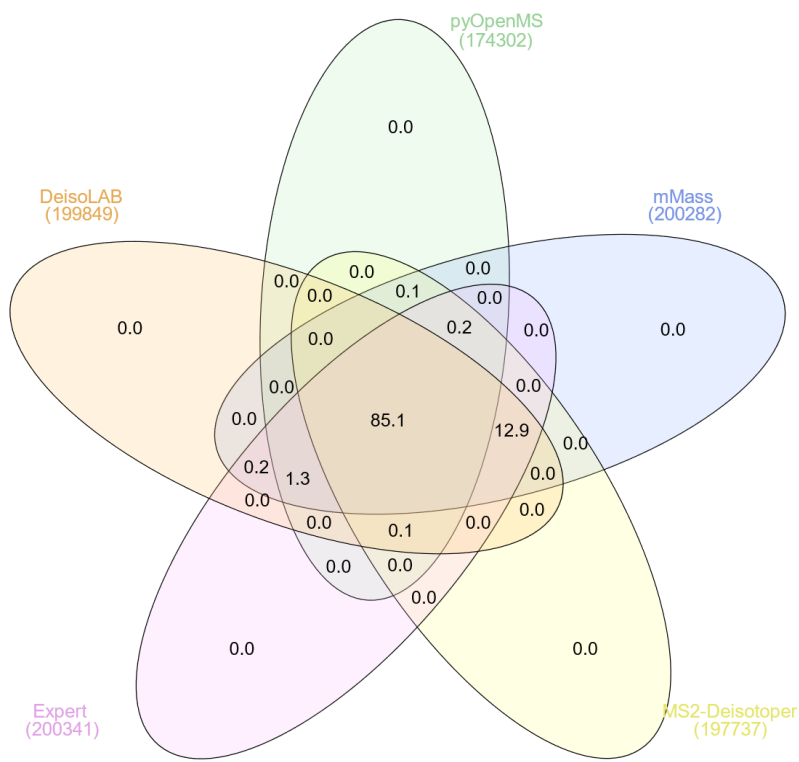

**Figure 35.** Venn diagram constructed using [55] – intersection between *nE* peaks identified by the expert, *DeisoLAB, mMass, pyOpenMS* and *MS2-Deisotoper*.

As it can be seen in *Figure 35*, the intersection is 85.1%. *mMass* slightly outperforms *DeisoLAB* for true negatives (TNs) but simultaneously reveals notably smaller values of true positives (TPs).

**Table 6.** Confusion matrix-based metrics for *DeisoLAB, mMass, pyOpenMS* and *MS2-Deisotoper*.

|  | DeisoLAB | mMass | MS2-Deisotoper | pyOpenMS |
|---|---|---|---|---|
| **TP** | 342 | 89 | 29 | 55 |
| **TN** | 199 828 | 200 008 | 197 403 | 173 994 |
| **FN** | 21 | 274 | 334 | 308 |
| **FP** | 513 | 333 | 2 938 | 26 347 |
| **Specificity [%]** | 99.74 | 99.83 | 98.53 | 86.85 |
| **Recall [%]** | 94.21 | 24.52 | 7.99 | 15.15 |
| **Balanced Accuracy [%]** | 96.98 | 62.18 | 53.26 | 51.00 |

As it can be noticed, the algorithm that reflects the number of accurately classified peaks belonging to an isotopic envelope the most is DeisoLAB. It is worth mentioning that nearly all of the investigated four algorithms have high negative accuracy (*Figure 36*), as the peaks not included in the isotopic envelope were correctly classified as the non-Envelope ones with specificity values ranging from 86.85% to 99.83% for all four algorithms (*Table 6*). Compared with DeisoLAB, the remaining three algorithms are characterised by significantly lower values of balanced accuracy (ranging from 51.00% to 96.98%), which provides information on how accurately the peaks included in isotopic envelopes and not included in any isotopic envelope were classified.

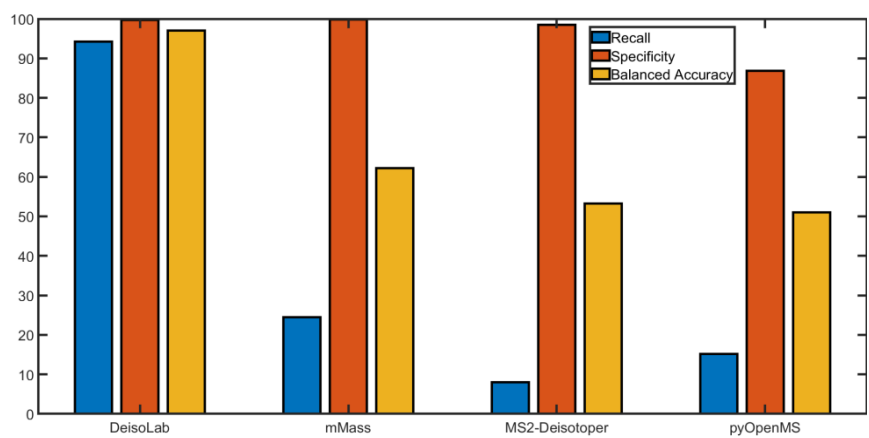

**Figure 36.** Confusion matrix metrix comparison for *DeisoLAB, mMass, pyOpenMS* and *MS2-Deisotoper*.

We conclude that according to all the aforementioned statistics measures, *DeisoLAB* exhibits better results, which is especially proven by comparing the recall values of all the methods.

## DISCUSSION

We conclude that the proposed algorithm for automatic identification of the isotopic envelope can handle large datasets from the MALDI type of experiments and the type of molecules to be analysed. The process of finding an isotopic envelope is divided into two parts: (1) a spectral similarity based on fuzzy logic and (2) a homogeneity based on spatial distribution; this makes the algorithm more robust and objective as it checks in two steps whether a pair of peaks is a member of an isotopic envelope or not. It is worth mentioning that the descriptors used in the classification evaluate the involvement of peak intensities in different ways, as they come from different groups (texture metrics, model component parameters).

The basic idea that a proper analysis of the spatial distribution of the molecular abundance in the tissue sample allows distinguishing between the envelope and non-envelope peaks has been proven, and the results are auspicious.

# ACKNOWLEDGEMENTS

We thank Prof. Jacek Łęski from Silesian University of Technology for comprehensive verification of fuzzy-inference systems and Katarzyna Frątczak for providing the pre-processed data. This work was co-financed by European Union grant under the European Social Fund, project no. POWR.03.02.00-00-I029.

# REFERENCES

[1] A. Misra, Ed., Challenges in delivery therapeutic genomics and proteomics, 1. ed., Amsterdam: Elsevier, 2011

[2] K.-H. Liang, Bioinformatics for biomedical science and clinical applications, 2013

[3] UPAC-IUB Joint Commission on Biochemical Nomenclature (JCBN) Nomenclature and Symbolism for Amino Acids and Peptides. Eur. J. Biochem. 1984, 138, 9–37

[4] Postolopoulos, V.; Bojarska, J.; Chai, T.-T.; Elnagdy, S.; Kaczmarek, K.; Matsoukas, J.; New, R.; Parang, K.; Lopez, O.P.; Parhiz, H.; et al. A Global Review on ShortPeptides: Frontiers and Perspectives. Molecules 2021, 26, 430

[5] D. L. Nelson, M. M. Cox and A. L. Lehninger, Eds., Principles of biochemistry, 4. ed., 5. print. ed., New York: Freeman, 2005

[6] M. Aichler and A. Walch, MALDI Imaging mass spectrometry: current frontiers and perspectives in pathology research and practice, vol. 95, pp. 422-431, 2015

[7] S. Shimma. Mass Spectrometry Imaging, vol. 11, pp. A0102-A0102, 2022

[8] J. L. Norris and R. M. Caprioli, "Analysis of Tissue Specimens by Matrix-Assisted Laser Desorption/Ionization Imaging Mass Spectrometry in Biological and Clinical Research," vol. 113, pp. 2309-2342, 2013

[9] P. Widłak, "Comparison of peptide cancer signatures identified by mass spectrometry in serum of patients with head and neck, lung and colorectal cancers: Association with tumor progression," International Journal of Oncology, September 2011

[10] Kurczyk, A.; Gawin, M.; Paul, P. Prognostic value of molecular intratumor heterogeneity in primary oral cancer and its lymph node metastases assessed by mass spectrometry imaging Molecules 2022, 27

[11] K. Bednarczyk, M. Gawin, M. Chekan, A. Kurczyk, G. Mrukwa, M. Pietrowska, J. Polanska and P. Widlak, "Discrimination of normal oral mucosa from oral cancer by mass spectrometry imaging of proteins and lipids," Journal of Molecular Histology, vol. 50, p. 1–10, November 2018

[12] A. Kurczyk, M. Gawin, M. Chekan, A. Wilk, K. Łakomiec, G. Mrukwa, K. Frątczak, J. Polanska, K. Fujarewicz, M. Pietrowska and P. Widlak, "Classification of Thyroid Tumors Based on Mass Spectrometry Imaging of Tissue Microarrays$\mathsemicolon$ a Single-Pixel Approach," International Journal of Molecular Sciences, vol. 21, p. 6289, August 2020

[13] A. Polanski, M. Marczyk, M. Pietrowska, P. Widlak and J. Polanska, "Signal Partitioning Algorithm for Highly Efficient Gaussian Mixture Modeling in Mass Spectrometry," PLOS ONE, vol. 10, p. e0134256, July 2015

[14] A. Polanski, M. Marczyk, M. Pietrowska, P. Widlak and J. Polanska, "Initializing the EM Algorithm for Univariate Gaussian, Multi-Component, Heteroscedastic Mixture Models by Dynamic Programming Partitions," vol. 15, p. 1850012, 2018

[15] N. Jaitly, A. Mayampurath, K. Littlefield, J. N. Adkins, G. A. Anderson and R. D. Smith, "Decon2LS: An open-source software package for automated processing and visualization

[16] A. Glodek and J. Polańska, "Method for mass spectrometry spectrum deisotoping based on fuzzy inference systems". Mathematica Applicanda, 46(1):77-86, 2018

[17] A. Glodek, "Fuzzy-inference system for isotopic envelope identification in Mass Spectrometry Imaging data". In: Rojas, I., Valenzuela, O., Rojas, F., Herrera, L.J., Ortuño, F. (eds) Bioinformatics and Biomedical Engineering. IWBBIO 2022. Lecture Notes in Computer Science, vol. 13347. Springer, Cham., 2022

[18] A. Polanski, M. Marczyk, M. Pietrowska, P. Widlak and J. Polanska, "Signal Partitioning Algorithm for Highly Efficient Gaussian Mixture Modeling in Mass Spectrometry," PLOS ONE, vol. 10, p. e0134256, July 2015

[19] A. Polanski, M. Marczyk, M. Pietrowska, P. Widlak and J. Polanska, "Initializing the EM Algorithm for Univariate Gaussian, Multi-Component, Heteroscedastic Mixture Models by Dynamic Programming Partitions," vol. 15, p. 1850012, 2018

[20] D. F. Findley, "Counterexamples to parsimony and BIC," Annals of the Institute of Statistical Mathematics, vol. 43, p. 505–514, September 1991

[21] A. C. Bovik, Ed., Handbook of image and video processing, 2. ed. ed., Amsterdam [u.a.]: Elsevier, 2005

[22] M. Petrou and P. G. Sevilla, Image Processing, 2006

[23] S. Singh, D. Srivastava and S. Agarwal, GLCM and its application in pattern recognition, 2017

[24] J. Chaki and N. Dey, Texture Feature Extraction Techniques for Image Recognition, 2020

[25] M. Hall-Beyer, "Practical guidelines for choosing GLCM textures to use in landscape classification tasks over a range of moderate spatial scales," International Journal of Remote Sensing, vol. 38, p. 1312–1338, January 2017

[26] Q. Zhao, C.-Z. Shi and L.-P. Luo, "Role of the texture features of images in the diagnosis of solitary pulmonary nodules in different sizes," Chin J Cancer Res, vol. 26, pp. 451-458, 2014

[27] C. D. Wijetunge, I. Saeed, B. A. Boughton, J. M. Spraggins, R. M. Caprioli, A. Bacic, U. Roessner and S. K. Halgamuge, "EXIMS: an improved data analysis pipeline based on a new peak picking method for EXploring Imaging Mass Spectrometry data," vol. 31, pp. 3198-3206, 2015

[28] G. H. John, R. Kohavi and K. Pfleger, Irrelevant Features and the Subset Selection Problem, 1994, pp. 121-129

[29] C. M. Bishop, Pattern recognition and machine learning, Corrected at 8th printing 2009 ed., New, York: Springer, 2009

[30] X. Liu, Y. Inbar, P. C. Dorrestein, C. Wynne, N. Edwards, P. Souda, J. P. Whitelegge, V. Bafna and P. A. Pevzner, "Deconvolution and Database Search of Complex Tandem Mass Spectra of Intact Proteins," vol. 9, pp. 2772-2782, 2010

[31] S. McIlwain, D. Page, E. L. Huttlin and M. R. Sussman, "Using dynamic programming to create isotopic distribution maps from mass spectra," vol. 23, pp. i328-i336, 2007

[32] J. Samuelsson, D. Dalevi, F. Levander and T. Rognvaldsson, "Modular, scriptable and automated analysis tools for high-throughput peptide mass fingerprinting," vol. 20, pp. 3628-3635, 2004

[33] P. Du and R. H. Angeletti, "Automatic Deconvolution of Isotope-Resolved Mass Spectra Using Variable Selection and Quantized Peptide Mass Distribution," vol. 78, pp. 3385-3392, 2006

[34] X.-j. Li, E. C. Yi, C. J. Kemp, H. Zhang and R. Aebersold, "A Software Suite for the Generation and Comparison of Peptide Arrays from Sets of Data Collected by Liquid Chromatography-Mass Spectrometry," vol. 4, pp. 1328-1340, 2005

[35] E. J. Breen, F. G. Hopwood, K. L. Williams and M. R. Wilkins, "Automatic Poisson peak harvesting for high throughput protein identification," vol. 21, pp. 2243-2251, 2000

[36] Y. Sun, J. Zhang, U. Braga-Neto and E. R. Dougherty, "BPDA - A Bayesian peptide detection algorithm for mass spectrometry," vol. 11, 2010

[37] J. F. de-Cossio, "Isotopica: a tool for the calculation and viewing of complex isotopic envelopes," vol. 32, pp. 3779-3779, 2004

[38] K. Xiao, F. Yu, H. Fang, B. Xue, Y. Liu and Z. Tian, "Accurate and Efficient Resolution of Overlapping Isotopic Envelopes in Protein Tandem Mass Spectra," Scientific Reports, vol. 5, October 2015

[39] Z. Yuan, J. Shi, W. Lin, B. Chen and F.-X. Wu, "Features-Based Deisotoping Method for Tandem Mass Spectra," Advances in Bioinformatics, vol. 2011, p. 1–12, January 2011

[40] X. Liu, Y. Inbar, P. C. Dorrestein, C. Wynne, N. Edwards, P. Souda, J. P. Whitelegge, V. Bafna and P. A. Pevzner, "Deconvolution and Database Search of Complex Tandem Mass Spectra of Intact Proteins," vol. 9, pp. 2772-2782, 2010

[41] A. P. Tay, A. Liang, J. J. Hamey, G. Hart-Smith and M. R. Wilkins, "MS2-Deisotoper: A Tool for Deisotoping High-Resolution MS/MS Spectra in Normal and Heavy Isotope-Labelled Samples," PROTEOMICS, vol. 19, p. 1800444, August 2019

[42] D. M. Horn, R. A. Zubarev and F. W. McLafferty, "Automated reduction and interpretation of," vol. 11, pp. 320-332, 2000

[43] N. Jaitly, A. Mayampurath, K. Littlefield, J. N. Adkins, G. A. Anderson and R. D. Smith, "Decon2LS: An open-source software package for automated processing and visualization of high resolution mass spectrometry data," vol. 10, 2009

[44] H. L. Röst, U. Schmitt, R. Aebersold and L. Malmström, "pyOpenMS: A Python-based interface to the OpenMS mass-spectrometry algorithm library," vol. 14, pp. 74-77, 2014

[45] B. Y. Renard, M. Kirchner, H. Steen, J. A. J. Steen and F. A. Hamprecht, "NITPICK: peak identification for mass spectrometry data.," BMC bioinformatics, vol. 9, p. 355, August 2008

[46] K. Jeong, J. Kim, M. Gaikwad, S. N. Hidayah, L. Heikaus, H. Schlüter and O. Kohlbacher, "FLASHDeconv: Ultrafast, High-Quality Feature Deconvolution for Top-Down Proteomics," vol. 10, pp. 213-218.e6, 2020

[47] M. Strohalm, D. Kavan, P. Novák, M. Volný and V. Havlíček, "A Cross-Platform Software Environment for Precise Analysis of Mass Spectrometric Data," Analytical Chemistry, vol. 82, p. 4648–4651, May 2010

[48] K. Xiao, F. Yu, H. Fang, B. Xue, Y. Liu and Z. Tian, "Accurate and Efficient Resolution of Overlapping Isotopic Envelopes in Protein Tandem Mass Spectra," Scientific Reports, vol. 5, October 2015

[49] K. Park, J. Y. Yoon, S. Lee, E. Paek, H. Park, H.-J. Jung and S.-W. Lee, "Isotopic Peak Intensity Ratio Based Algorithm for Determination of Isotopic Clusters and Monoisotopic Masses of Polypeptides from High-Resolution Mass Spectrometric Data," vol. 80, pp. 7294-7303, 2008

[50] V. Zabrouskov, M. W. Senko, Y. Du, R. D. Leduc and N. L. Kelleher, "New and automated MS approaches for top-down identification of modified proteins," vol. 16, pp. 2027-2038, 2005

[51] Z. Zhang and A. G. Marshall, "A universal algorithm for fast and automated charge state deconvolution of electrospray mass-to-charge ratio spectra," vol. 9, pp. 225-233, 1998

[52] A. M. Mayampurath, N. Jaitly, S. O. Purvine, M. E. Monroe, K. J. Auberry, J. N. Adkins and R. D. Smith, "DeconMSn: a software tool for accurate parent ion monoisotopic mass determination for tandem mass spectra," vol. 24, pp. 1021-1023, 2008

[53] M. W. Senko, S. C. Beu and F. W. McLaffertycor, "Determination of monoisotopic masses and ion populations for large biomolecules from resolved isotopic distributions," vol. 6, pp. 229-233, 1995

[54] E. Scifo, G. Calza, M. Fuhrmann, R. Soliymani, M. Baumann and M. Lalowski, "Recent advances in applying mass spectrometry and systems biology to determine brain dynamics," Expert Review of Proteomics, vol. 14, p. 545–559, June 2017

[55] H. Heberle, G. V. Meirelles, F. R. da Silva, G. P. Telles and R. Minghim, "InteractiVenn: a web-based tool for the analysis of sets through Venn diagrams," vol. 16, 2015